\pdfoutput=1

\documentclass[11pt]{article}

\usepackage[]{emnlp2023-latex/acl}
\usepackage{comment}
\usepackage{graphicx}
\usepackage{times}
\usepackage{latexsym}
\usepackage{placeins}

\usepackage{booktabs}
\usepackage{caption} 
\usepackage{stfloats} 
\usepackage[T1]{fontenc}

\usepackage[utf8]{inputenc}

\usepackage{microtype}

\usepackage{inconsolata}

\usepackage[normalem]{ulem}

\title{A Study on Scaling Up Multilingual News Framing Analysis 
}

\author{Syeda Sabrina Akter and Antonios Anastasopoulos\\
  Department of Computer Science, George Mason University \\
  \texttt{\{sakter6,antonis\}@gmu.edu}}

\begin{document}
\maketitle
\FloatBarrier
\begin{abstract}
Media framing is the study of strategically selecting and presenting specific aspects of political issues to shape public opinion. Despite its relevance to almost all societies around the world, research has been limited due to the lack of available datasets and other resources. This study explores the possibility of dataset creation through crowdsourcing, utilizing non-expert annotators to develop training corpora. 
We 
first extend framing analysis beyond English news to a multilingual context (12 typologically diverse languages) through automatic translation. We also present a novel benchmark in Bengali and Portuguese on the immigration and same-sex marriage domains.
Additionally, we show that a system trained on our crowd-sourced dataset, combined with other existing ones, leads to 
a 5.32 percentage point increase from the baseline, showing that crowdsourcing is a viable option.
Last, we study the performance of large language models (LLMs) for this task, finding that task-specific fine-tuning is a better approach than employing bigger non-specialized models.\footnote{Code and Dataset available here: \url{https://github.com/syedasabrina/Scaling-up-multilingual-framing-analysis.git}}

\end{abstract}

\section{Introduction}
News framing refers to the power of the news media to define and interpret events, issues, and policies by emphasizing certain aspects while downplaying or excluding others. According to \citet{entman1993framing}, it can ``make a piece of information more noticeable, meaningful, or memorable to audiences''. 
It plays a crucial role in influencing how people interpret and react to information presented in news articles. The language used in news media can shape public opinion and reveal biases and agendas, which can ultimately shape the way people understand and react to current events.

\begin{figure}[h]
\centering
\includegraphics[width=0.5\textwidth]{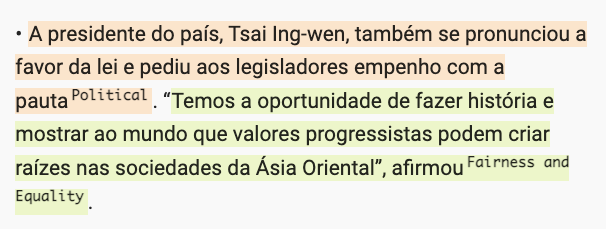}
\caption{The image illustrates the process of framing in Portuguese at the sentence level, showcasing how specific language for each sentence strategically shape a Political and Equality narrative in the same article.}
\label{fig:img1}
\vspace{-1.5em}
\end{figure}

Traditionally, framing analysis has relied on manual annotation by linguists, social studies experts, and trained annotators, lacking the potential of AI-driven systems leading to a rather limited explorations of automating framing analysis. Moreover, existing studies have been restricted primarily to English-only data, leaving a gap in research concerning multilingual and low-resource contexts.

Our work focuses on employing NLP techniques for the framing analysis task to automate the analysis process, extract insights from large datasets efficiently, and identify patterns in the language used in news media. To address these challenges, \citet{boydstun2014tracking} introduced a codebook, Policy Frames Codebook, based on which the Media Frames Corpus \cite[\texttt{MFC};][]{DBLP:conf/acl/CardBGRS15} was created. This dataset is comprised broad categories of common policy frames and annotations of US news articles. However, the availability of such datasets in languages beyond English remains limited. 

Getting a higher volume of higher quality data (such as, \texttt{MFC}) is time and resource intensive. Hence, we study the alternative of gathering a high volume of comparatively lower quality but easy-to-collect data. We achieve this through crowdsourcing and automatic translation techniques. We also examine the combination of lower and  higher quality data.


In this study, we first \textbf{introduce a new crowd-sourced dataset: Student-sourced Noisy Frames Corpus (SNFC)}. We have achieved time and cost efficiency by involving a large number of semi-trained annotators for the data collection and annotation process of the corpus. \texttt{SNFC} covers immigration and same-sex marriage domains and includes \textbf{novel benchmark test sets in Bengali and Portuguese}, offering new perspectives in these languages. Additionally, we automatically expand multilinguality to the task by translating the \texttt{MFC} and \texttt{SNFC} to 12 more languages. We show that a neural classifier trained on the combination of both \texttt{MFC} and \texttt{SNFC} yields significant performance improvements, both in English as well as in a multilingual setting. Finally, we explore generative large language models, such as LLaMA \cite{touvron2023llama}, to study their efficacy  for this task.

Our findings show that neural models trained on \texttt{SNFC} can reach the performance levels of those trained on high quality data (i.e., \texttt{MFC}). Going further, we find that the combination of expert and non-expert annotated data (i.e. \texttt{MaSNFC+MFC}) outperforms just MFC, which provides a path towards expanding coverage without the need for expensive expert annotations.


\section{Related Work}
\label{sec:related}


Framing analysis provides valuable insights into different perspectives on news topics across various countries and languages. However, there is a notable lack of research and annotated corpora for framing analysis in languages other than English. This limitation hinders our understanding of media framing in different parts of the world and other societies' opinion regarding specific issues. 
To address this gap, a multilingual approach is essential in analyzing media framing across diverse linguistic and cultural contexts. \citet{DBLP:conf/emnlp/AliH22} provide a comprehensive survey of the framing analysis task, focusing specifically on studies in English datasets exploring various approaches and techniques employed in framing analysis.

Two prominent datasets used for framing analysis are the Media Frames Corpus \cite[\texttt{MFC};][]{DBLP:conf/acl/CardBGRS15} and the Gun Violence Frames Corpus \cite[GVFC;][]{liu-etal-2019-detecting}. The \texttt{MFC}, annotated according to the guidelines provided in the codebook of \citet{boydstun2014tracking}, covers 6 different political issues including immigration, same-sex marriage, and gun violence, among others. It includes both article headlines and news texts, providing a broader and more comprehensive dataset. On the other hand, the GVFC focuses solely on the topic of gun violence, with 10 manually annotated frames defined in a different codebook, and it only includes article headlines.

\citet{DBLP:conf/acl/AkyurekGEIBW20} extended the GVFC by curating headlines in German, Turkish, and Arabic following the same process as the original dataset from the respective news websites, specifically targeting keywords related to gun violence and mass shootings. The frames used in the multilingual datasets remained consistent with those in the GVFC, and is the one of the few multilingual sources for this task. Additionally, the Australian Parliamentary Speeches (APS) dataset~\cite{DBLP:conf/acl-alta/KhanehzarTM19} offers another perspective on framing analysis, as it consists of transcripts speeches related to same-sex marriage bills presented in the Australian Parliament. Although the APS dataset focuses on data 
from a country other than the United States, it is still limited to English language texts, which narrows the scope of the framing analysis task.

The \texttt{MFC} has served as a valuable resource in various framing-related studies. For example, it was used to develop a semi-supervised model by extracting a Russian lexicon from their Russian test corpora which consists of news articles sourced from reputable Russian newspapers~\cite{DBLP:conf/emnlp/FieldKWPJT18}. 
In a different vein, \citet{naderi-hirst-2017-classifying} used it to benchmark sentence-level classification tasks, employing LSTM, BiLSTM, and GRU-based systems. Considering the significant contributions of this corpus to the field, we have incorporated it into our system for training and evaluation purposes, alongside our \texttt{SNFC} dataset.

Several studies have employed various techniques such as topic modeling \cite{dimaggio2013exploiting,roberts2014structural,nguyen2015guided}, cluster analysis \cite{burscher2016frames}, and neural networks~\cite{naderi-hirst-2017-classifying,DBLP:conf/acl-alta/KhanehzarTM19,mendelsohn-etal-2021-modeling,DBLP:conf/websci/KwakAA20} to construct systems for framing analysis. These investigations have consistently demonstrated that leveraging state-of-the-art pre-trained models based on transformers~\cite{DBLP:conf/naacl/DevlinCLT19,zhuang-etal-2021-robustly,DBLP:conf/acl/ConneauKGCWGGOZ20} is a highly effective approach, yielding significantly improved results compared to other techniques. In our study, we follow the state of the art and build models similar to those employed by \citet{liu-etal-2019-detecting} and \citet{DBLP:conf/acl-alta/KhanehzarTM19}.

We also investigated crowdsourcing methods which, as defined by~\citet{howe2006rise}, is an online, distributed problem-solving and production model that leverages the collective intelligence of online communities for specific goals. This technique aims to tap into the global talent pool, accelerating innovation and problem-solving across various domains. ~\citet{hossain2015crowdsourcing} provide a comprehensive literature review, identifying numerous crowdsourcing methods, which emphasizes the difficulty of generalizing these methods due to their diversity and application-specific nature. However, the widespread use of these methods demonstrates versatility and adaptability of different crowdsourcing methods. ~\citet{zhao2014evaluation} suggest that future research should focus on standardizing crowdsourcing processes to enhance efficiency and effectiveness. This indicates an increasing realization of the necessity to codify crowdsourcing approaches, notwithstanding their inherent variability.

\section{Dataset Creation}
\label{sec:dataset}
\FloatBarrier
In this section, we present our methodology for curating \texttt{SNFC} training dataset through crowdsourcing (\S\ref{sec:snfc}) and outline the process of extending the dataset to incorporate multilinguality (\S\ref{sec:multilingual}). Lastly, we introduce our innovative Portuguese and Bengali benchmarks, highlighting their significance in the context of this study (\S\ref{sec:noveltest}).

\paragraph{\texttt{SNFC} Training Corpus}
\label{sec:snfc}

To construct the crowd-sourced training portion of the \texttt{SNFC}, we turned to students at George Mason University. In particular, this was done as part of an in-class assignment for a graduate-level natural language processing class with about 80 students involved.\footnote{We are releasing these data with the students' consent.}

The students were presented with the challenge of building a Media Frames Analysis system (effectively, a sentence-level neural classifier), without having access to significant amounts of data. In particular, the students were provided only with a description of the codebook of \citet{boydstun2014tracking} presented in Table~\ref{table:codebooks}, along with 250 sentence-level examples called the seed dataset from the \texttt{MFC} corpus sampled so that all 15 frame dimensions were present.

The codebook and the samples were meant to facilitate the annotators' understanding of the task. The only other information available to them was that their final systems would be evaluated on multiple languages (see \S\ref{sec:multilingual}) on the immigration and same-sex marriage domains.\footnote{These evaluation sets were based on the \texttt{MFC} test sets.} 

The students were first tasked with procuring 150 new sentences each, from any source and in any language, and label them, according to the codebook, to be used as their ``first'' training set. They then had to produce an additional 150 sentences which would then be annotated by two of their peers (so that we will be able to measure inter-annotator agreement). Any label disagreements were resolved by the students, by obtaining an additional label for majority voting. All in all, each student produced a minimum of 300 annotated sentences. While the students had the option to collect data in any language, all of them, apart from two, collected and annotated the initial data in English. The two other students who collected data in different languages chose their native languages: Telugu, and Hindi.



To collect the data, the students were allowed to do anything they wanted. They ended up utilizing diverse techniques that range from targeted web scraping to generating sentences with the assistance of AI tools such as, ChatGPT~\cite{radford2019language}. 
We can broadly categorize the sources of data into three categories: AI tools (such as ChatGPT and ChatSonic), online news platforms (including Online Articles, NBC, CNN, BBC, and NYTimes), and social media platforms (such as Twitter and Reddit). Students have used a combination of two or more categories to collect their data. Around 77\% of students used AI tools, 14.8\% relied on social media platforms, and 67.9\% used online news platforms for data collection purposes. It is important to note that, AI was only used by the students in the first step of data collection. This shows how artificial intelligence (AI) eases the process of collecting relevant, topic-specific text. The process of data validation and labeling was entirely done by human annotators.


In the end, we ended up with a total of 17,520 sentences from the combined student training corpus of 300 sentences each, eliminating the occasional duplicate instances. The dataset has a generally substantial inter-annotator agreement, with a Cohen's $\kappa$ \cite{cohen1960coefficient} coefficient of~0.61. 

To further contextualize this, we note that the inter-annotator agreement of the \texttt{MFC} (as detailed in the paper) is assessed using Krippendorff's $\alpha$ \cite{krippendorff2011computing}, with respective values of 0.08 and 0.20 for the domains of same-sex marriage and immigration. \texttt{SNFC} (our dataset) combines sentences from both of these domains and the Krippendorff's $\alpha$ value for \texttt{SNFC} stands at 0.103 which is similar to the one of \texttt{MFC}. Given that this is a 15-way classification task, we believe the inter-annotator agreement for \texttt{SNFC} is not particularly low for such a nuanced task.




\paragraph{Multilinguality}
\label{sec:multilingual}
\FloatBarrier
To benchmark media framing beyond English our first step is to simply translate the original \texttt{MFC} dataset into other languages. We use machine translation\footnote{Google Translate, specifically.} to translate all sentences of the \texttt{MFC} corpus into 12 typologically diverse languages, namely Bengali, German, Greek, Italian, Turkish, Nepali, Hindi, Portuguese, Telugu, Russian, Swahili, and Mandarin Chinese. 

While the primary reason for this process is the ability to benchmark the task on other languages (as well as the inability to collect annotated test sets in all of these languages -- see also \S\ref{sec:noveltest}), this simple data augmentation technique is also a reasonable way to also obtain training data in other languages. Hence, we perform this translation both on the training and the dev/test portions of the dataset, and combine all languages to form the multilingual version of the dataset.

\begin{table}[t]
\centering
\begin{tabular}{rc}
\toprule
\textbf{Language Pair} & \textbf{Rating (\%)}\\
\midrule
 English-Bengali & 61.2  \\
 English-Greek & 73.4  \\
 English-Hindi & 77.4  \\
 English-Nepali & 47.2  \\
 \midrule
 Comet Score (All languages) & 76.05 \\
\bottomrule
\end{tabular}
\caption{Average rating for Human Evaluation of the Automatic Translation Quality}
\label{tab:rating}
\vspace{-2em}

\end{table}

Lastly, the same translation models were used to augment our crowd-sourced \texttt{SNFC} dataset to cover all of the above-mentioned languages.


We have studied the quality of the translation through human assessment. For each language, we took 100 translations from English and had them reviewed by bilingual speakers who scored the translations on a scale from 1 to 10 based on accuracy and clarity. For this evaluation, we used four languages: Bengali, Greek, Hindi, and Nepali. From the average rating for each language pair (See Table~\ref{tab:rating}), we observe that the average rating is higher for higher resourced languages like Greek and Hindi. On the other hand, Nepali, being the only lower resourced language, has a lower rating of 4.72 out of 10, suggesting that perhaps Nepali results should be taken with a grain of salt, as the reason for general poor performance is likely to be the low quality of the translations. 

We have also further performed quality estimation over all translations by calculating the \texttt{CometKiwi} score~\cite{rei2023scaling} of the translations. Note that we resort to automatic quality estimation since we do not have access to reference translations. The overall score of 76.05\% is in line with our human evaluation over the sample, and suggests that automatic translations are largely reliable in our dataset. The higher scores for the high resource languages of the human-evaluation and \texttt{CometKiwi} (see Appendix~\ref{sec:appendix3} for a breakdown by language) indicate that automatic translations can be a reasonable alternative to gathering large quantities of high quality multilingual data for the framing task.

\paragraph{Novel Test Set}
\label{sec:noveltest}

\begin{figure*}[t]
\centering
\vspace{-1em}
\includegraphics[width=0.90\textwidth]{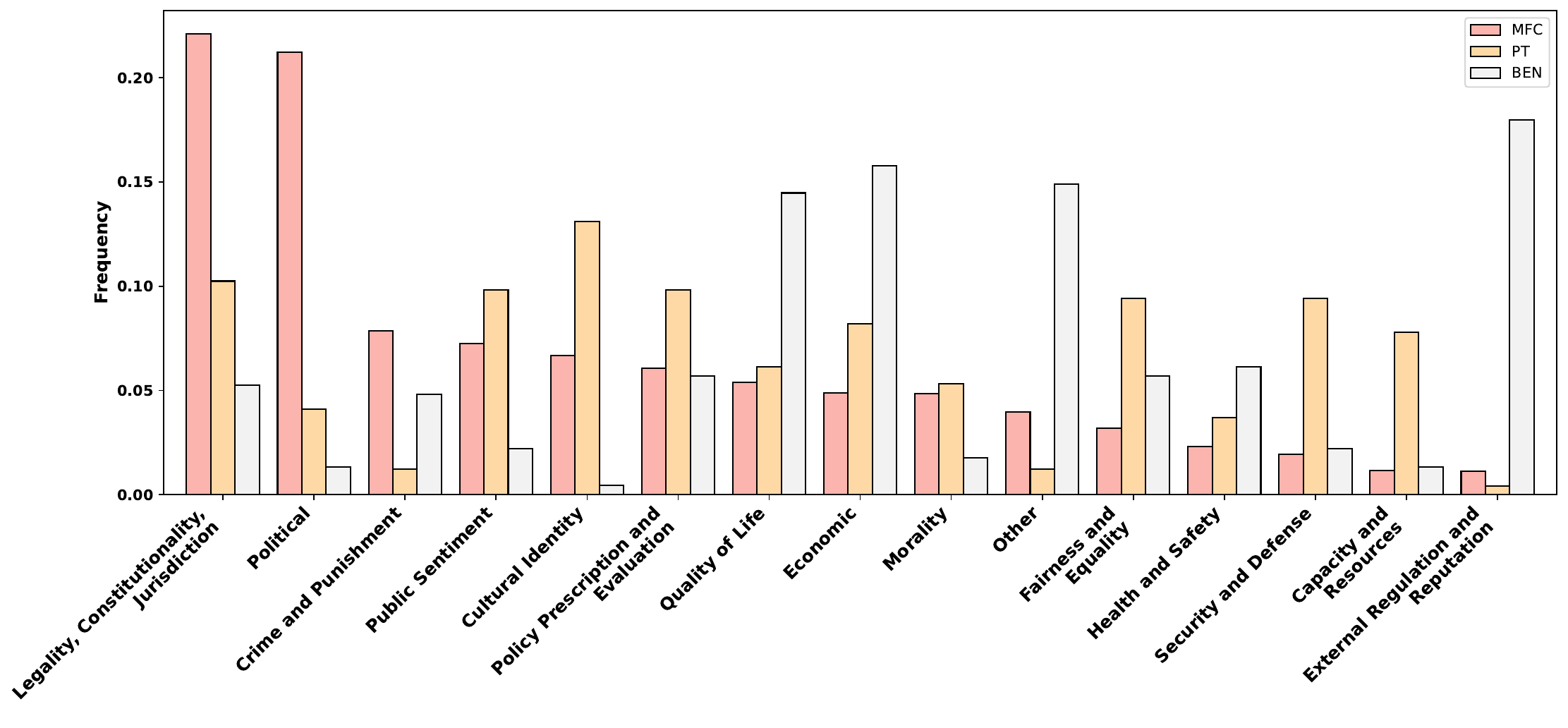}
\caption{The label distributions of the \texttt{MFC} and our new Bengali and Portuguese test sets. Note that they differ significantly.}
\label{fig:img7}
\label{tab:num_per_frame}
\vspace{-1em}
\end{figure*}


While the automatic translation of the \texttt{MFC} benchmark is a reasonable start for our multilingual exploration, it does not come without drawbacks: the provided text, regardless of the language, is only relevant to the USA cultural context.

To even better benchmark the quality of framing analysis systems on different language and cultural contexts, we create a pair of novel test sets in (Bangladesh) Bengali and (Brazilian) Portuguese. 
The news articles used in this test set were sourced from reputable newspapers in Bangladesh and Brazil, aligning with the chosen domains of immigration and same-sex marriage. 

Each test set is comprised of~of 10 news articles for each language. The annotators were native speakers of the languages and they adhered closely to the definitions provided by the authors (Table~\ref{table:codebooks}), ensuring consistency with the labels found in the \texttt{MFC}. 



Figure~\ref{fig:img7} shows the label distribution for the \texttt{MFC} and the novel test set, listing the number of sentences per frame in each language. In the case of Bengali, the news articles predominantly focus on the immigration domain, reflecting the cultural disparities between Brazil and Bangladesh. Specifically, the test set emphasizes the economic and lifestyle aspects of immigration (Bengali), while also delving into the legal and policy-making dimensions of the domain (Portuguese).

It is of note that the two benchmarks, despite being rather small, still show interesting differences in terms of their label distribution. For example, the most common label on the Bengali set is "External Regulation and Reputation", which is the least common one in the Portuguese one. And the reverse is the case for the "Cultural Identity" label which is the most common in Portuguese and least common in Bengali. Another interesting observation is that the Bengali test set contains more data labeled as "Other" compared to the other two languages. Upon analyzing the data with the help of a native speaker, we found that most of the Bangladeshi articles emphasize a lot on reporting information in the form of dates and numbers, rather than offering opinions on the issues.


\section{Framing Analysis System and Results}

\paragraph{Experimental Setup}

We approach the task as a multilabel classification problem \cite{tsoumakas2007multi}, leveraging the pretrained RoBERTa~\cite{zhuang-etal-2021-robustly} language model, similar to the SOTA approach employed by \citet{DBLP:conf/acl-alta/KhanehzarTM19}. For all models we set the maximum sequence length to 256, with a batch size of 16,
and train using a learning rate of $10^{-5}$. To expand to more languages, we employ the multilingual XLM-RoBERTa model \cite{DBLP:conf/acl/ConneauKGCWGGOZ20}. Throughout all experiments, we use the base model size.\footnote{Appendix~\ref{tab:eng_acc2} and \ref{tab:multi_acc2} also provides results with the BERT and mBERT~\cite{DBLP:conf/naacl/DevlinCLT19} models (but RoBERTa and XLM-R consistently outperformed BERT and mBERT.}

We first report results with models exclusively trained on \texttt{MFC}, and \texttt{SNFC} datasets, as well as their concatenation.
To investigate a more data-scarce scenario, we also compiled a smaller sample consisting of about 10\% of the original \texttt{MFC}, named \texttt{MFC10}, ensuring all 15 target labels are included. 
Beyond the single-dataset baselines, we combine the expert-annotated \texttt{MFC} and \texttt{MFC10} with our crowd-sourced \texttt{SNFC}. To further study the effect of the size of the \texttt{SNFC}, we have experimented with \texttt{SNFC50}, a randomized halved subset of the original \texttt{SNFC} that is more closer to the \texttt{MFC} in size.


\paragraph{English Results and Discussion}

We first establish the usefulness of our crowdsourced data, by focusing on the performance on the original  test set of the English \texttt{MFC} dataset (using the monolingual RoBERTa model).
Results are presented in Table~\ref{tab:eng_acc}. 


\begin{table}[t]
\centering
\begin{tabular}{p{2.2cm}ll}
\toprule
\textbf{Tr. Data} & \textbf{\#Sentences}  & \textbf{Accuracy} \\
\midrule
\multicolumn{3}{@{}l}{Baselines}\\
\texttt{MFC} & 9739 &  69.52 \\
\texttt{MFC10} & 1125 &  57.45 \\
\midrule
\multicolumn{3}{@{}l}{including crowd-sourced data}\\
\texttt{SNFC} & 17520 &  54.37 \\
\texttt{SNFC50} & 8760 & 54.7  \\
\texttt{MFC+SNFC}  & 27260 &  72.07 \\
\texttt{MFC+SNFC50}  & 18499 &  72.89 \\
\texttt{MFC10+SNFC} & 18645 &  64.75 \\
\texttt{MFC10+SNFC50} & 9885 &  62.05 \\
\midrule
\multicolumn{3}{@{}l}{filtered crowd-sourced data}\\
\texttt{MaSNFC} & 5182 &  48.77 \\
\texttt{MFC+MaSNFC}  & 14922 &  \textbf{73.22} \\
\texttt{MFC10+MaSNFC} & 6307  & 60.94 \\
\bottomrule
\end{tabular}
\caption{Mean Accuracy Scores on the \texttt{MFC} evaluation set for RoBERTa models trained on English Datasets. \# stands for "number of".} 
\label{tab:eng_acc}
\vspace{-2em}
\end{table}

First, it is worth pointing out that relying solely on crowd-sourced data is not promising: the \texttt{SNFC}-only training underperforms both the \texttt{MFC}-only setting, as well as the \texttt{MFC10}-only setting, which has only around 10\% of the training data size!

However, combining the expert-annotated data with the crowd-sourced ones yields significant improvements over the expert-only baselines, as \texttt{MFC+SNFC} yields an extra 2.5 accuracy points over \texttt{MFC} (72\% vs 69.5\%). The improvement is even larger (more than 7 accuracy points) in the resource-restricted \texttt{MFC10} scenario. The accuracy remains consistent both with \texttt{SNFC50} alone and when combined with \texttt{MFC}, as \texttt{MFC+SNFC50} and \texttt{MFC+SNFC} yield similar results, indicating that performance gains are not merely due to larger data volume.

\paragraph{Filtering of Crowdsourced Data}
Given the potential for noise in any crowd-sourced dataset, we explore a simple filtering technique to sample more high-quality crowd-sourced. In particular, we obtain sentence-level representations for each sentence, and select only the \texttt{SNFC} instances that exhibit more than 85\% cosine similarity with any \texttt{MFC} instance. Effectively, we select \texttt{SNFC} sentences that are most similar to \texttt{MFC} ones. We refer to this sample as \texttt{MFC}-aligned \texttt{SNFC} (\texttt{MaSNFC}).

Results with this (almost 3x smaller) sample are more encouraging (Table~\ref{tab:eng_acc}): combining \texttt{MaSNFC} with \texttt{MFC} yields our best model with an accuracy of 73.22. In the data-scarce scenario of \texttt{MFC10}, adding \texttt{MaSNFC} is again beneficial, but including the whole unfiltered \texttt{SNFC} is even better.

These findings underline the promise of crowd-sourcing for collecting a high volume of (somewhat) lower quality data. The performance improvement for the \texttt{MaSNFC+MFC} shows promise for the combination of low-volume high-quality along with a higher-volume of lower-quality data. This approach effectively balances the depth and breadth of the dataset, leveraging the strengths of both data types.


\paragraph{Multilingual Results and Discussion}

For the first part of our multilingual experiments, we employ a translate-train and translate-test scenario. All of the dataset samples introduced above were translated to all 12 evaluation languages, and we now replicate the same experimental setups as above, the only difference being that we will use a multilingual LM (XLM-R instead of RoBERTa).
All results are presented in Table~\ref{tab:multi_acc} (which presents the average accuracy across the 12 languages for \texttt{mMFC}, as well as performance on our novel Bengali and Portuguese benchmark). 

\begin{table}[t]
\centering
\begin{tabular}{@{ }p{2.1cm}l|c@{ }c@{}}
\toprule
\textbf{Tr. Data} & {\textbf{\texttt{mMFC}}} & {\small{\textbf{\textsc{Bengali}}}} & {\small{\textbf{\textsc{Portuguese}}}} \\
\midrule
\multicolumn{4}{@{}l}{Zero-shot (only English train)} \\
\small{\texttt{MFC}}  & 28.13 & 25.44 & 28.28    \\
\multicolumn{4}{@{}l}{Baselines (translate-train)} \\
\small{\texttt{MFC}} &  44.99 &  25.88 &  \textbf{33.61}   \\
\small{\texttt{MFC10}} &  28.64 &  23.68 &  27.87  \\
\midrule
\multicolumn{4}{@{}l}{+ crowd-sourced (translate-train)} \\
\small{\texttt{SNFC}} &  28.04 & 25.44 & 23.77    \\
\small{\texttt{MFC+SNFC}}  & 44.07 &  26.31 &  31.56  \\
\small{\texttt{MFC10+SNFC}} &  33.11 & \textbf{32.02} & 26.62      \\
\midrule
\multicolumn{4}{@{}l}{+ filtered crowd-sourced (translate-train)} \\
\small{\texttt{MaSNFC}} & 27.55 &  16.67 & 15.98  \\
\small{\texttt{MFC+MaSNFC}}  &  \textbf{45.73} &  28.07 &  \textbf{33.61}  \\
\small{\texttt{MFC10+MaSNFC}} &  32.56 &  24.56 & 26.64  \\
\bottomrule
\end{tabular}
\caption{Mean Accuracy Scores on the \texttt{MFC} evaluation set and Novel Multilingual Test Set for XLM-R models trained on Multilingual Datasets. The best scores have been highlighted.}
\label{tab:multi_acc}
\vspace{-1em}
\end{table}

\begin{figure*}[t]
\centering
\vspace{-1em}
\includegraphics[scale=0.5]{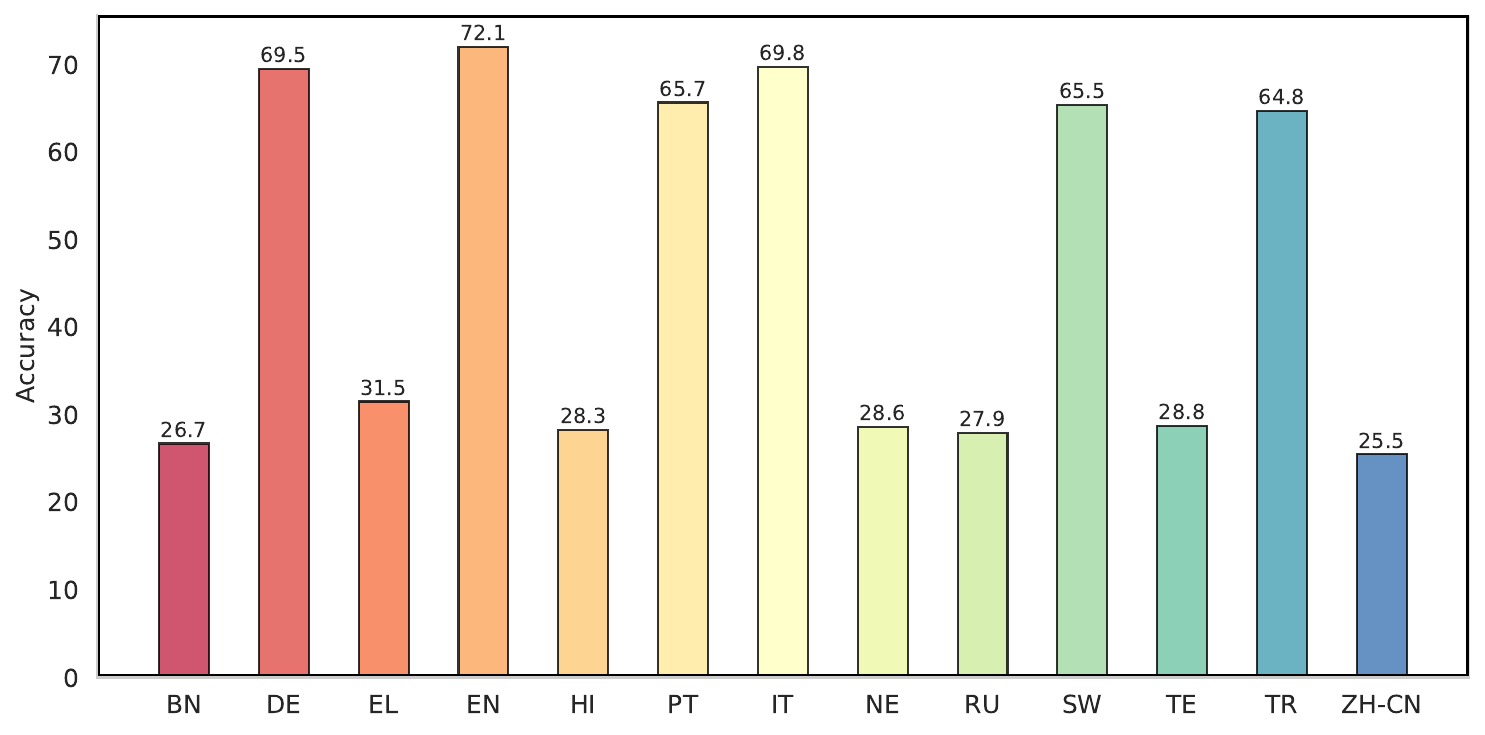}
\vspace{-1em}
\caption{The best model performs very inequitably across languages on \texttt{mMFC}. The highest accuracy is in English (72.1\%) followed by Italian and German, while other languages from non-western countries (e.g. Bengali, Hindi, Chinese, and others) have much lower performance (under 30\%).}
\label{fig:img2}
\end{figure*}

First of all, we show that relying on zero-shot cross-lingual transfer, without employing the translate-train technique is not a competitive baseline.
The translated \texttt{MFC} baseline is competitive on average, but as we discuss below it performs quite inequitably across languages. As before, combining expert annotated data with filtered crowd-sourced ones (\texttt{MFC+MaSNFC}) is best. Our findings from the monolingual experiments generally hold in the multilingual ones.

In the Bengali test set, the inclusion of all crowd-soured data improves upon the baseline by a small margin. The improvement from filtered crowdsourced data is more modest. However, it is interesting that the best performance is obtained when using fewer expert annotations (\texttt{MFC10+SNFC}), improving by almost 6 percentage points over the baseline! We hypothesize that using the whole \texttt{MFC} dataset overfits the US context -- but we leave this analysis for future work.
In the Portuguese test set, we observe generally similar patterns as in the \texttt{mMFC}, with the exception that we do not observe any improvement from the crowd-sourced data. We leave a further investigation for future work.

We note that the accuracies for the Bengali and Portuguese test sets are significantly lower than those of the English \texttt{MFC} and the \texttt{mMFC} test sets. We suspect that the training data, being automatic translations, may not capture the nuances of the original news articles. Second, the domain shift due to cultural context differences between training and test may play a significant role. To improve the scores further, it may be necessary to obtain original news articles from diverse culturally distinct sources in different languages.

\paragraph{\texttt{mMFC} Breakdown per Language}

We further analyse the per-language performance of our best-performing model on \texttt{mMFC} (see Figure~\ref{fig:img2}). English accuracy (72.1) is en par with the monolingual setting (73.2), and German, Italian, Swedish, and Turkish also yield accuracies higher than 64\%. But for other languages the model performs much worse, including high-resource ones like Greek (31.5\%), Russian (28\%), and Chinese (25.5\%). While translation errors may play a role here, we are confident that they are not enough to explain such a large discrepancy. For example, while Nepali has admittedly low-quality translations (see previous discussion), Hindi, Greek, and Chinese certainly have translations of fairly high quality and yet they fall in the same low performance ballpark.
We suspect that this gap may only be bridged through data collection (either expert- or crowd-annotated) in the appropriate languages and cultural contexts.

\begin{figure*}[t]
\centering
\vspace{-1em}
\includegraphics[width=\textwidth]{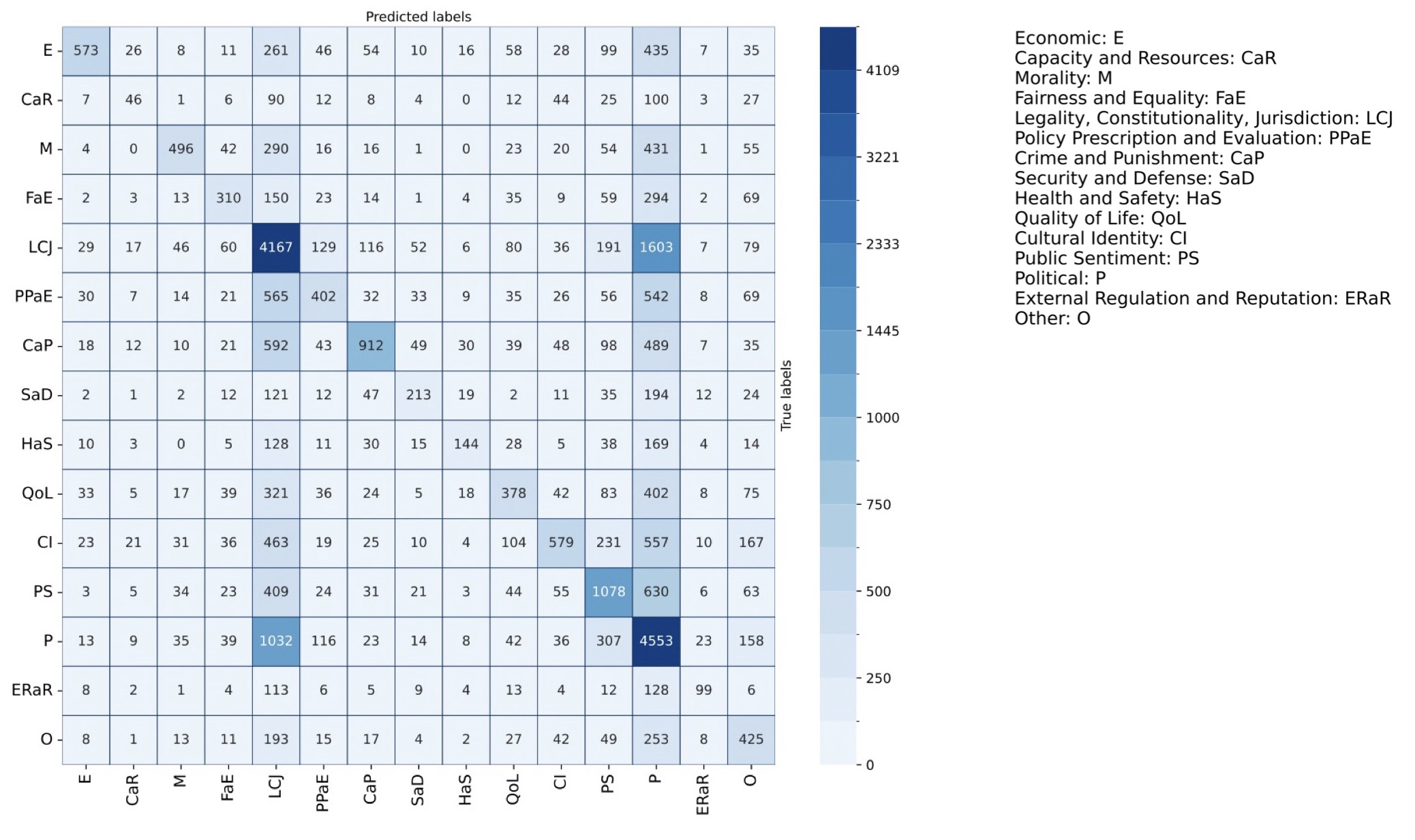}
\caption{Confusion matrix for the best model's prediction for the mMFC Test set.}
\label{fig:heatmap}
\vspace{-1.5em}
\end{figure*}

\paragraph{Error Analysis}
We analyzed the errors using a confusion matrix for our best-performing model \texttt{MFC+MaSNFC} on the \texttt{mMFC} evaluation set, as shown in Figure~\ref{fig:heatmap}. The heat-map reveals that out of 15 labels, 9 achieve the majority of instances correctly. Specifically, the labels `Political' and `Legality, Constitutionality, Jurisdiction' have the highest number of instances predicted correctly. However, when the model makes incorrect predictions, the errors are mainly categorized into the `Political' and `Legality, Constitutionality, Jurisdiction' labels. This led us to suspect a potential data imbalance in our training model. Further examination of the data confirmed that these two labels indeed have a majority of instances in the training set, leading to the tendency to predict these labels when uncertain. 

One could also further argue that these two labels are quite close semantically and hence their confusion is perhaps expected.
We have examined the original data from MFC for the immigration and same-sex issues, which were used to train our baseline model. This dataset indeed shows a skewed distribution with a disproportionate number of instances falling under these two labels. This suggests that US-based news articles covering these domains inherently tend to fall in these two categories. Given the domain, we deduce that such an imbalance in label distribution might be a common trend in news articles from other countries as well. This assumption can be further validated in our novel test sets derived from Bangladesh and Brazil, which also reveal a similar inclination towards certain labels, as discussed in the previous section.

\section{Generative Language Models}
\label{sec:llms}

LLMs like GPT-4 \cite{OpenAI2023GPT4TR}, Falcon \cite{penedo2023refinedweb}, and LLaMA \cite{touvron2023llama}, are trained on vast amounts of text and have shown immense promise in a variety of NLP tasks. Their broad knowledge base qualifies them as potential tools for framing analysis. In this study, we have also explored three of these models, particularly the open-sourced ones: Mistral, LLaMA-2, and Falcon.  

\begin{table}[t]
\centering
\begin{tabular}{rc}
\toprule
\textbf{Model} & \textbf{Accuracy (\%)}\\
\midrule
 Falcon-40b-instruct & 22.95  \\
 \textbf{Mistral-7B-Instruct-v0.1} & \textbf{35.33}  \\
 Llama2-chat-70B & 22.22  \\
\bottomrule
\end{tabular}
\caption{Exact Match accuracy of the LLMs. The highest accuracy (35\%, \textbf{bolded}) is significantly worse than the task finetuned RoBERTa model's performance (73.22\%).}
\label{tab:llms_score}
\vspace{-1em}
\end{table}

\paragraph{Experimental Setting}

The instruction presents the framing task as a multiple choice question with 15 options and we have curated the instruction to include the definitions of all the labels, similar to the ones the students have used to annotate the \texttt{SNFC}. The instruction we use is given in Appendix~\ref{sec:appendix4}. We conduct all experiments in the zero shot setting, to assess the potential of LLMs to generalize and apply their knowledge effectively without task-specific training. The experiments were run on the English only test set (\texttt{MFC-test}) to ensure comparability with other task-finetuned models previously evaluated on the same test set.

\paragraph{Results and Discussion}

The results (see Table~\ref{tab:llms_score}) show the exact match accuracy of different LLMs on the \texttt{MFC-test} dataset. The performance of Llama2-chat-70B aligns closely with that of Falcon-40b-instruct, and Mistral-7B-Instruct-v0.1 outperformed them significantly showing that the sheer size of a model does not necessarily equate to better performance.

Interestingly, the best performance was achieved by employing smaller, task-finetuned models, with RoBERTa achieving an exact match accuracy of 73.22\%. This significantly surpasses the highest result for general LLMs, as their best performance is at 35.33\%, observed with Mistral-7B-Instruct-v0.1. This difference in performance highlights the importance of task-specific fine-tuning on model efficacy. The finetuning process allows models like RoBERTa to adapt their parameters more closely to the nuances of the specific task, resulting in a more precise understanding and response generation compared to models that rely solely on broad, generalized training. The results also suggests that there is a trade-off between model size and specialized training. While larger models have a vast knowledge base, they are not always effective in applying this knowledge to specific tasks without fine-tuning. 

\paragraph{Error Analysis}

The LLMs exhibit a range of errors in predicting the correct frames for the provided texts (See Table~\ref{tab:model_predictions}). These errors include spelling mistakes, overgeneralization, assigning multiple labels where only one is appropriate, and misinterpretation. Generally, the models struggle with adhering to instructions, such as inventing new frames rather than selecting from the provided list (External Regulatory and Renown). Additionally, a common issue among all three of the models is their failure to introduce their answers concisely as instructed. Contrary to the clear direction to reply only with the label name, they begin responses with phrases like `The most suitable frame is...'. 

The Mistral 7B model achieves a higher accuracy rate compared to the other two model; however, it often adds additional commentary to its responses. The LLaMA-2 70B model's predictions are inconsistent, notably when it replaces `External Regulation and Reputation' with `External Regulatory and Renown', demonstrating a tendency towards misrepresentation. The Falcon 40B sometimes accurately identifies the frame but fails to use the exact label name, responding with `Economical' instead of `Economic'.

Since the models have the tendency to predict labels with spelling errors and synonymous labels, we have employed different techniques to measure the accuracy of these models to ensure a true reflection of the system's performance. To derive the correct label names from synonymous words and to overlook spelling mistakes, we employed the FastText~\cite{joulin2016fasttext} and Edit Distance~\cite{levenshtein1966binary} algorithms. These were used to determine the textual similarity between the models' predictions and the 15 labels they were intended to predict.

\section{Conclusion}

In conclusion, our study emphasizes the importance of data quality and language diversity in multilingual framing analysis. Combining the Media Frames Corpus (\texttt{MFC}) with the Student-Sourced Noisy Frames Corpus (\texttt{SNFC}) yields significant improvements, highlighting the value of  larger datasets despite the annotation quality potentially being lower. However, lower accuracy in multilingual experiments indicates the need for accurate translations and culturally diverse training data to improve multilingual framing analysis. Last, the sub-par performance of LLMs showcases a future research direction towards task-specific finetuning of the LLMs.

\section*{Limitations}

The main limitation of this study is that it relies on automated translation via Google Translator to introduce multilinguality to the task. It is well known that the translations conducted by Google Translator may not achieve the same level of quality as authentic translations. Moreover, for lower-resource languages such as Nepali and Swahili, the translations obtained from Google Translator may not fully capture the nuances and characteristics as well as it probably can if translated to higher-resource languages as German or Greek. Additionally, since the \texttt{MFC} dataset primarily consists of US news sources, the translations into different languages does not adequately reflect the biases and perspectives surrounding a specific political issue in different countries. 
We attempt to mitigate this limitation with our new Bengali and Portuguese test sets.
Collecting more data from different countries in different languages will eventually address this limitation, but we leave this large-scale undertaking for the future.

\section*{Acknowledgements}
We are thankful to the anonymous reviewers for their useful feedback, as well as to the students of the GMU CS 678 course and the annotators for the Bengali and Portuguese test set  who majorly contributed in the creation of our crowdsourced dataset.
This project was supported by the National Science Foundation under grant IIS-2327143.
This project was also supported by resources provided by the Office of Research Computing at George Mason University (\url{https://orc.gmu.edu}) and funded in part by grants from the National Science Foundation (Awards Number 1625039 and 2018631).

\bibliography{custom}

\clearpage
\newpage

\appendix

\section{Annotation Schema}

We used a crowdsourcing approach with the help of non-expert annotators to create our training corpus, simplifying the process compared to the traditional method of hand-annotating by expert linguists and social science scholars, which is both expensive and inefficient. We collected data for the corpus in collaboration with graduate students whose task was to gather 150 sentences each, in various languages, from news articles related to the domains of immigration and same-sex marriage.
These sentences were then annotated using the 15 framing dimensions established in the study \cite{boydstun2014tracking}, which are globally accepted, shown in Table \ref{table:codebooks}.

\vspace*{-2em}
\begin{table*}[!b]
\begin{tabular}{p{4.5cm} p{10cm}}
\toprule
\textbf{Frames}                                        & \textbf{Definitions}                    \\                               
\midrule
Economic                                         & The financial consequences and economic implications of the matter on various levels (person, family, community or broader economy).      \\ 
Capacity and Resources                        & The presence or absence of various resources(physical, geographic, human, and financial) and the ability of existing systems.             \\ 
Morality                                      & Perspectives, policy objectives, or actions driven by religious principles, duties, ethics, or social responsibilities.                   \\
Fairness and Equality                         & The balance or distribution of laws, rights, and resources among individuals or groups.                                                   \\
Legality, Constitutionality, Jurisdiction & Discusses rights, freedoms and authority of individuals, corporations, and government.                                                    \\
Policy Prescription and Evaluation            & Specific policies proposed to address identified issues and the assessment of policy effectiveness.                                       \\
Crime and Punishment                             & Effectiveness and implications of laws and their enforcement.                                                                             \\
Security and Defense                          & Actions or calls to action aimed at protecting individuals, groups, or nations from potential threats to their well-being.                \\
Health and Safety                             & Access to healthcare, health outcomes, disease, sanitation, mental health, violence prevention, infrastructure safety, and public health. \\
Quality of life                               & Threats and opportunities for the individual's wealth, happiness and well being.                                                          \\
Cultural Identity                             & Traditions, customs or values of a social group in relation to a policy issue.                                                            \\
Public Sentiment                                & References of attitudes and opinions of the general public, including polling and demographics.                                           \\
Political                                     & Political considerations, actions, efforts, stances, and partisan, bipartisan, or lobbying activities related to an issue.                \\
External Regulation and Reputation          & The external relations of nations or groups, trade agreements, policy outcomes, and external perceptions or consequences.                 \\
Other                                         & Frames that don't fit into the categories above.        \\                                                                           
\bottomrule
\end{tabular}
\captionsetup{justification=centering}  
\caption{Frames and their definitions as outlined by Policy Frames Codebook (PFC, \citet{boydstun2014tracking}). This codebook was given to the students as annotation schema.}
\label{table:codebooks}
\end{table*}

\clearpage

\section{Novel Bengali and Portuguese Test Set Statistic}
\label{sec:appendix1}

\begin{table}[h]
\centering
\begin{tabular}{p{3.7cm} p{1cm} p{1.6cm} }
\toprule
\textbf{Number of sentences} & \textbf{Bengali} & \textbf{Portuguese}\\
\midrule
 Economic & 36 & 20 \\
 Capacity and Resources & 3 & 19 \\
 Morality & 4 & 13 \\
 Fairness and Equality & 13 & 23 \\
 Legality Constitutionality Jurisdiction & 12 & 25 \\
 Policy Prescription and Evaluation & 13 & 24 \\
 Crime and Punishment & 11 & 3 \\
 Security and Defence & 5 & 23 \\
 Health and Safety & 14 & 9 \\
 Quality of Life & 33 & 15 \\
 Cultural Identity & 1 & 32 \\
 Public Sentiment & 5 & 24 \\
 Political & 3 & 10 \\
 External Regulation and Reputation & 41 & 1 \\
 Other & 34 & 3 \\
\midrule
Total & 228 & 244 \\
\bottomrule
\end{tabular}
\caption{Number of texts per frame per language}
\label{tab:num_per_frame}
\end{table}

The distribution of labels in the Bengali and Portuguese test sets (see Table \ref{tab:num_per_frame}) reveals intriguing domain affinity. In the case of Bengali, the news articles predominantly focus on the immigration domain, reflecting the cultural disparities between Brazil and Bangladesh. Specifically, the test set emphasizes the economic and lifestyle aspects of immigration (Bengali), while also delving into the legal and policy-making dimensions of the domain (Portuguese).

\clearpage

\section{Assessing Translation Quality}
\label{sec:appendix3}

Table~\ref{tab:comet} shows the breakdown of the comet score per language. 

\begin{table}[h]
\centering
\begin{tabular}{p{4cm} p{1.6cm}}
\toprule
\textbf{Language Pair} & \textbf{Comet Score (\%)}\\
\midrule
 English-Bengali & 74.39  \\
 English-German & 76.93  \\
 English-Greek & 76.64  \\
 English-Hindi & 67.87  \\
 English-Italian & 79.04  \\
 English-Nepali & 86.84  \\
 English-Russian & 79.87  \\
 English-Swahili & 73.71  \\
 English-Telugu & 69.02  \\
 English-Bengali & 78.79  \\
 English-Turkish & 74.63  \\
 English-Chinese & 74.63  \\
 English-Portuguese & 74.89  \\
 \midrule
 System Score & 76.05 \\
\bottomrule
\end{tabular}
\caption{Average score from \texttt{CometWiki} of the Automatic Translation Quality without reference. The high resource languages (i.e., Italian, Greek etc) have higher scores than lower resource languages (i.e., Telugu)}
\label{tab:comet}
\end{table}

\clearpage

\section{Complete Results for English and Multilingual Experiments}
\label{sec:appendix2}

We observed the mean accuracy of the \texttt{MFC} evaluation set for models trained on English and Mulitlingual datasets. The key findings are summarized below:

\begin{enumerate}
    \item The \texttt{MFC} alone achieved higher accuracy compared to other systems, with scores of 61.93\% and 69.52\% for BERT and RoBERTa-based models, respectively. However, when using the \texttt{MFC10} dataset with limited high-quality data, the accuracy dropped significantly to 53.02\% and 57.45\% for BERT and RoBERTa models, respectively.
    \item The \texttt{SNFC} and \texttt{MaSNFC} datasets exhibited lower accuracy when evaluated individually, compared to the \texttt{MFC}. However, the \texttt{SNFC} outperformed \texttt{MFC10} in terms of accuracy for the BERT model. The \texttt{SNFC} has an accuracy of 60.57\% while the \texttt{MFC10} has gotten 53.02\%. It is worth noting that the larger size of the \texttt{SNFC} contributed to its higher accuracy compared to \texttt{MaSNFC}, which is almost three times smaller.
    \item Combining the \texttt{MFC} with our datasets led to substantial accuracy improvements. The models trained on \texttt{MFC+SNFC}  (72.57\%, 72.07\%) and \texttt{MFC+MaSNFC}  (72.85\%, 73.22\%) achieved higher accuracy than the \texttt{MFC} alone (61.93\%, 69.52\%), for both BERT and RoBERTa models.
    \item Combining \texttt{\texttt{MFC}10} with our datasets, we observed improved accuracy as well. The \texttt{MFC10+SNFC} combination yielded an accuracy improvement of 6.1 and 4.77 percentage points for BERT and RoBERTa models, respectively, compared to \texttt{MFC10}. Similarly, \texttt{MFC10+MaSNFC} demonstrated a similar improvement of 7.1 and 3.49 percentage points, respectively.
    \item The overall accuracies of the \texttt{MFC} evaluation set for multilingual data (Table 3) are lower compared to the accuracies for English training (Table 2). This can be attributed to the fact that the training data in other languages were obtained through automatic translation, which may not be of the same quality as human translations or original news articles in those languages.
    \item Among the datasets, \texttt{MFC+MaSNFC} achieved the highest accuracy of 45.73 on the multilingual test set, outperforming both \texttt{MFC} and \texttt{MFC10} datasets.
    \item For the Bengali test set, the highest accuracy (32.02) was achieved by the \texttt{MFC10+SNFC} training dataset. As for the Portuguese test set, the highest accuracy of 33.61 was obtained by two systems: \texttt{MFC} and \texttt{MFC+MaSNFC}.
    \item Overall, the accuracies for the Bengali and Portuguese test sets were lower than those for the \texttt{MFC} evaluation set. This can be attributed to two factors. First, the training data, being translations, may not capture the nuances of the original news articles. Second, the training data mainly consists of \texttt{MFC}, which is collected from US-based news media sources. The test sets, on the other hand, were collected from Brazil and Bangladesh, which have different cultural contexts in their news articles that cannot be fully replicated through translation. To improve the scores further, it would be necessary to obtain original news articles from diverse culturally distinct sources in different languages.
\end{enumerate}

\begin{table}[t]
\centering
\begin{tabular}{p{1.5cm}p{1.8cm}p{1.1cm}p{1.5cm}}
\hline
\textbf{System Name} & \textbf{Number of Sentences} & \textbf{BERT} & \textbf{RoBERTa} \\
\hline
\texttt{MFC} & 9740 & 61.93 & 69.52 \\
\texttt{MFC10} & 1125 & 53.02 & 57.45 \\
\texttt{SNFC} & 17520 & 60.57 & 54.37 \\
\texttt{MaSNFC} & 5182 & 52.05 & 48.77 \\
\texttt{MFC+ SNFC}  & 27260 & 72.57 & 72.07 \\
\texttt{MFC+ MaSNFC}  & 14922 & \textbf{72.85} & \textbf{73.22} \\
\texttt{MFC10+ SNFC} & 18645 & 68.03 & 64.75 \\
\texttt{MFC10+ MaSNFC} & 6307 & 60.12 & 60.94 \\
\hline
\end{tabular}
\caption{Mean Accuracy Scores on the \texttt{MFC} evaluation set for models trained on English Datasets. The best scores have been highlighted.}
\label{tab:eng_acc2}
\end{table}

\begin{table*}[t]
\centering
\begin{tabular}{ p{3cm}|p{1.5cm}p{1.5cm}|p{1.5cm}p{1.5cm}|p{1.5cm}p{1.5cm}}
\hline
\textbf{System Name} & \multicolumn{2}{|c|}{\textbf{\texttt{MFC} Evaluation Set}} & \multicolumn{2}{|c|}{\textbf{Bengali Test Set}} & \multicolumn{2}{|c}{\textbf{Portuguese Test Set}} \\
 & mBERT & XLM-R & mBERT & XLM-R & mBERT & XLM-R \\
\hline
\texttt{MFC} (English) & 27.70 & 28.13 & 16.67 & 25.44 & 26.23 & 28.28   \\
\texttt{MFC} & 44.87 & 44.99 & 21.93 & 25.88 & 30.33 & \textbf{33.61}   \\
\texttt{MFC10} & 27.7 & 28.64 & 20.61 & 23.68 & 30.33 & 27.87  \\
\texttt{SNFC} & 28.05 & 28.04 & 22.37 & 25.44 & 27.05 & 23.77  \\
\texttt{MaSNFC} & 28.86 & 27.55 & 11.84 & 16.67 & 20.49 & 15.98  \\
\texttt{MFC+SNFC}  & \textbf{45.09} & 44.07 & \textbf{23.25} & 26.31 & 29.92 & 31.56  \\
\texttt{MFC+MaSNFC}  & 44.42 & \textbf{45.73} & 22.37 & 28.07 & \textbf{31.97} & \textbf{33.61}  \\
\texttt{MFC10} + \texttt{SNFC} & 30.01 & 33.11 & 25 & \textbf{32.02} & 29.51 & 26.62  \\
\texttt{MFC10+MaSNFC} & 33.33 & 32.56 & 22.81 & 24.56 & 22.13 & 26.64  \\
\hline
\end{tabular}
\caption{Mean Accuracy Scores on the \texttt{MFC} evaluation set and Novel Multilingual Test Set for models trained on Multilingual Datasets. The best scores have been highlighted.}
\label{tab:multi_acc2}
\end{table*}

The study highlights challenges in multilingual framing analysis, with lower accuracies compared to English training. It emphasizes the need for high-quality translations and original news articles. Combining datasets like \texttt{MFC+MaSNFC} can enhance accuracy. Considering cultural and linguistic contexts and diverse training data is crucial for better understanding framing across languages and cultures.

\clearpage

\section{Instruction for the Generative AI Models}
\label{sec:appendix4}

This was the instruction that was given to the models discussed in Section~\ref{sec:llms}.
\\

"In this task, you will be provided with a list of frames and a sentence. Your goal is to select the single most suitable frame from the given list for the provided sentence. Frames are cognitive structures that help humans interpret information by providing a mental framework for understanding. Each frame represents a specific perspective, context, or interpretation. Frame Selection Format: In your response, do not write anything other than the name of the frame. Frames List and Definitions:{'Economic': 'The financial consequences and economic implications of the matter on various levels (person, family, community or broader economy).','External Regulation and Reputation': 'The external relations of nations or groups, trade agreements, policy outcomes, and external perceptions or consequences.','Political': 'Political considerations, actions, efforts, stances, and partisan, bipartisan, or lobbying activities related to an issue.','Public Sentiment': 'References of attitudes and opinions of the general public, including polling and demographics.','Cultural Identity': 'Traditions, customs, or values of a social group in relation to a policy issue.', 'Quality of Life': 'Threats and opportunities for the individual's wealth, happiness, and well-being.','Health and Safety': 'Access to healthcare, health outcomes, disease, sanitation, mental health, violence prevention, infrastructure safety, and public health.','Security and Defense': 'Actions or calls to action aimed at protecting individuals, groups, or nations from potential threats to their well-being.','Crime and Punishment': 'Effectiveness and implications of laws and their enforcement.','Policy Prescription and Evaluation': 'Specific policies proposed to address identified issues and the assessment of policy effectiveness.','Legality, Constitutionality, Jurisdiction': 'Discusses rights, freedoms, and authority of individuals, corporations, and government.','Fairness and Equality': 'The balance or distribution of laws, rights, and resources among individuals or groups.','Morality': 'Perspectives, policy objectives, or actions driven by religious principles, duties, ethics, or social responsibilities.','Capacity and Resources': 'The presence or absence of various resources (physical, geographic, human, and financial) and the ability of existing systems.','Other': 'Frames that don't fit into the categories above.'}   Please select the most appropriate frame for the given sentence, and specify the chosen frame without additional commentary. You are allowed to choose only one frame from the provided list that best aligns with the meaning and context of the given sentence. Consider the nuances of the sentence and the various frames in the list before making your selection. Your choice should reflect the frame that you believe most accurately captures the intended interpretation of the sentence. Remember that frames can significantly impact how information is perceived and understood. Choose the frame that enhances the sentence's meaning and effectively shapes the interpretation. Your ability to accurately match sentences with appropriate frames will contribute to successful communication and understanding. Do not write anything like, 'The selected frame is.' Input: "

\section{Erroneous Predictions from Generative LLMs}

\begin{table*}[ht!]
\centering
\begin{tabular}{p{7cm}|p{2cm}|p{2cm}|p{3cm}}
\toprule
\textbf{Text} & \textbf{Gold Label Output} & \textbf{Models} & \textbf{Model Prediction} \\
\midrule
DETENTION OF REFUGEES IN TEXAS OK'D & Crime and Punishment & Llama 70B & The External Regulations And reputation was picked since it refers nation borders control migrants entering its territory as refugee status; \\
\midrule
But last spring, the attorney for the regents, Burns Newsome, advised them to end the waivers for illegal immigrants, saying the practice is at odds with SB 529. Now, the regents are having a series of public hearings to get input on the issue. & Legality, Constitutionality, Jurisdiction & Llama 70B & External Regulatory and Renown \\
\midrule
3 Prison Guards Guilty of Abuse Of Immigrants & Crime and Punishment & Falcon 40B & Prison guards guilty. \\
\midrule
DHS budget brinkmanship could leave local governments in the lurch & Economic & Falcon 40B & Economical \\
\midrule
Disorder On The Border & Security and Defense & Mistral 7B & The most suitable frame for the sentence 'Disorder on the Border' is 'Security and Defense' \\
\midrule
Police say Orleans-Lindsay shot Toatley, 37, to death Oct. 30 during an undercover drug purchase. Orleans-Lindsay, of Silver Spring, allegedly accepted \$3,500 in cash for cocaine, pretended to get the drugs and returned to the trooper's sport-utility vehicle a few minutes later, firing a single shot. Toatley, a 16-year trooper and father of three, died 2 1/2 hours later. & Crime and Punishment & Mistral 7B & 'Law Enforcement', 'Political', 'Fairness and Equality', 'Crime and Punishment' and 'Legality, Constitutionality, Jurisdiction' \\
\bottomrule
\end{tabular}
\caption{LLMs predictions compared to gold label outputs. Errors include spelling errors to not following instruction.}
\label{tab:model_predictions}
\end{table*}

\end{document}